\RequirePackage{fix-cm}
\documentclass[twocolumn,natbib]{svjour3}          
\smartqed  
\usepackage{graphicx}
\usepackage{mathptmx}      
\usepackage{color,xcolor}
\usepackage{epsfig}
\usepackage{graphicx}

\usepackage{adjustbox}
\usepackage{array}
\usepackage{booktabs}
\usepackage{colortbl}
\usepackage{float,wrapfig}
\usepackage{hhline}
\usepackage{multirow}
\usepackage{subcaption} 
\usepackage{makecell}

\usepackage{amsfonts}
\usepackage{mathtools}
\usepackage{bm}
\usepackage{nicefrac}

\usepackage{changepage}
\usepackage{extramarks}
\usepackage{fancyhdr}
\usepackage{lastpage}
\usepackage{setspace}
\usepackage{soul}
\usepackage{xspace}

\usepackage[pagebackref=true,breaklinks=true,colorlinks,citecolor=blue,bookmarks=false]{hyperref}
\usepackage{url}

\usepackage{algorithm, algorithmic}
\usepackage{enumerate}

\newcolumntype{L}[1]{>{\raggedright\let\newline\\\arraybackslash\hspace{0pt}}m{#1}}
\newcolumntype{C}[1]{>{\centering\let\newline\\\arraybackslash\hspace{0pt}}m{#1}}
\newcolumntype{R}[1]{>{\raggedleft\let\newline\\\arraybackslash\hspace{0pt}}m{#1}}

\newcommand{\sect}[1]{Section~\ref{#1}}

\newcommand{\fig}[1]{Figure~\ref{#1}}
\newcommand{\tbl}[1]{Table~\ref{#1}}

\newcommand{\ignorethis}[1]{}

\def\data{Vimeo-90K\xspace}
\def\model{TOFlow\xspace}
\def\toflow{TOFlow\xspace}
\def\fflow{Fixed Flow\xspace}
\def\epicflow{EpicFlow\xspace}
\def\vbm4d{V-BM4D\xspace}
\def\denoisingbenchmark{Vimeo denoising/deblocking benchmark\xspace}
\def\srbenchmark{Vimeo super-resolution benchmark\xspace}
\def\interpbenchmark{Vimeo interpolation benchmark\xspace}

\def\flownet{motion estimation\xspace}
\def\procnet{video processing\xspace}

\makeatletter
\DeclareRobustCommand\onedot{\futurelet\@let@token\@onedot}
\def\@onedot{\ifx\@let@token.\else.\null\fi\xspace}

\def\eg{e.g\onedot} 
\def\ie{i.e\onedot}

\makeatother

\definecolor{MyDarkBlue}{rgb}{0,0.08,1}
\definecolor{MyDarkGreen}{rgb}{0.02,0.6,0.02}
\definecolor{MyDarkRed}{rgb}{0.8,0.02,0.02}
\definecolor{MyDarkOrange}{rgb}{0.40,0.2,0.02}
\definecolor{MyPurple}{RGB}{111,0,255}
\definecolor{MyRed}{rgb}{1.0,0.0,0.0}
\definecolor{MyGold}{rgb}{0.75,0.6,0.12}
\definecolor{MyDarkgray}{rgb}{0.66, 0.66, 0.66}

\newcommand{\myparagraph}[1]{\vspace{5pt}\noindent\textbf{#1}}

\journalname{International Journal of Computer Vision}
\begin{document}

\title{Video Enhancement with Task-Oriented Flow}

\titlerunning{International Journal of Computer Vision}

\author{Tianfan Xue$^1$ \and
        Baian Chen$^2$ \and
        Jiajun Wu$^2$ \and
        Donglai Wei$^3$ \and
        William T. Freeman$^{2,4}$
}

\authorrunning{International Journal of Computer Vision}

\institute{Tianfan Xue \at
           \email{tianfan@google.com}
           \and
           Baian Chen \at
           \email{baian@mit.edu}
           \and
           Jiajun Wu \at
           \email{jiajunwu@mit.edu}
           \and
           Donglai Wei \at
           \email{donglai@seas.harvard.edu}
           \and
           William T. Freeman \at
           \email{billf@mit.edu}
           \and
           $^1$\;\; Google Research, Mountain View, CA, USA \\
           $^2$\;\; Massachusetts Institute of Technology, Cambridge, MA, USA \\
           $^3$\;\; Harvard University, Cambridge, MA, USA \\
           $^4$\;\; Google Research, Cambridge, MA, USA
}

\date{Received: date / Accepted: date}

\maketitle

\begin{abstract}

Many video enhancement algorithms rely on optical flow to register frames in a video sequence. Precise flow estimation is however intractable; and optical flow itself is often a sub-optimal representation for particular video processing tasks. In this paper, we propose task-oriented flow (\model), a motion representation learned in a self-supervised, task-specific manner. We design a neural network with a trainable motion estimation component and a video processing component, and train them jointly to learn the task-oriented flow. For evaluation, we build \data, a large-scale, high-quality video dataset for low-level video processing. \model outperforms traditional optical flow on standard benchmarks as well as our \data dataset in three video processing tasks: frame interpolation, video denoising/deblocking, and video super-resolution. 

\end{abstract}

\section{Introduction}
\label{sec:intro}

\begin{figure*}[t]
    \centering
    \includegraphics[width=\textwidth]{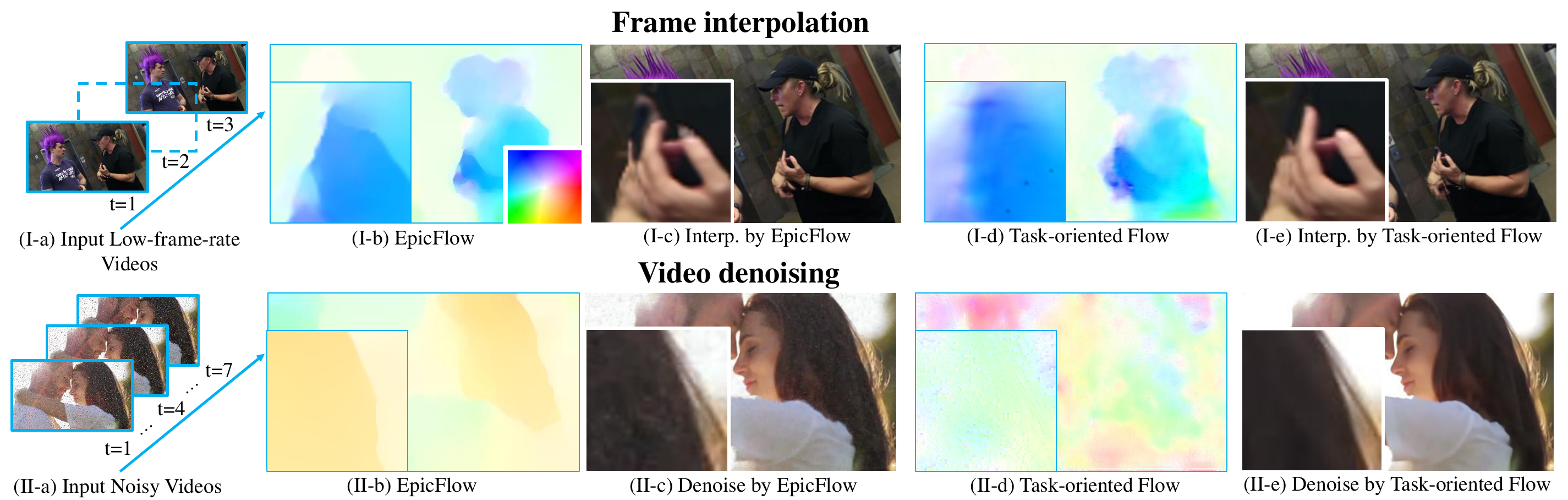}
    \caption{
    Many video processing tasks, \eg, temporal frame-interpolation (top) and video denoising (bottom), rely on flow estimation. In many cases, however, precise optical flow estimation is intractable and could be suboptimal for a specific task. For example, although EpicFlow~\citep{Revaud2015Epicflow} predicts precise movement of objects (I-b, the flow field aligns well with object boundaries), small errors in estimated flow fields result in obvious artifacts in interpolated frames, like the obscure fingers in (I-c). With the task-oriented flow proposed in this work (I-d), those interpolation artifacts disappear as in (I-e). Similarly, in video denoising, our task-oriented flow (II-d) deviates from EpicFlow (II-b), but leads to a cleaner output frame (II-e). Flow visualization is based on the color wheel shown on the corner of (I-b).
    }
    \label{fig:teaser} 
\end{figure*}

Motion estimation is a key component in video processing tasks such as temporal frame interpolation, video denoising, and video super-resolution. Most motion-based video processing algorithms use a two-step approach~\citep{liu2011bayesian,baker2011database,liu2010high}: they first estimate motion between input frames for frame registration, and then process the registered frames to generate the final output. Therefore, the accuracy of flow estimation greatly affects the performance of these two-step approaches.

However, precise flow estimation can be challenging and slow for many video enhancement tasks. The brightness constancy assumption, which many motion estimation algorithms rely on, may fail due to variations in lighting and pose, as well as the presence of motion blur and occlusion. Also, many motion estimation algorithms involve solving a large-scale optimization problem, making it inefficient for real-time applications. 

Moreover, solving for a motion field that matches objects in motion may be sub-optimal for video processing. \fig{fig:teaser} shows an example in frame interpolation. \epicflow \citep{Revaud2015Epicflow}, one of the state-of-the-art motion estimation algorithms, calculates a precise motion field (I-b) whose boundary is well-aligned with the fingers in the image (I-c); however, the interpolated frame (I-c) based on it still contains obvious artifacts due to occlusion. This is because \epicflow only matches the visible parts between the two frames; however, for interpolation we also need to inpaint the occluded regions, where \epicflow cannot help. In contrast, task-oriented flow, which we will soon introduce, learns to handle occlusions well (I-e), though its estimated motion field (I-d) differs from the ground truth optical flow. 
Similarly, in video denoising, \epicflow can only estimate the movement of the girl's hair (II-b), but our task-oriented flow (II-d) can remove the noise in the input. 
Therefore, the frame denoised by ours is much cleaner than that by \epicflow (II-e). For specific video processing tasks, there exist motion representations that do not match the actual object movement, but lead to better results.

In this paper, we propose to learn this task-oriented flow (\model) representation with an end-to-end trainable convolutional network that performs motion analysis and video processing simultaneously.
Our network consists of three modules: the first estimates the motion fields between input frames; the second registers all input frames based on estimated motion fields; and the third generates target output from registered frames.
These three modules are jointly trained to minimize the loss between output frames and ground truth. Unlike other flow estimation networks~\citep{Ranjan2017Optical,Fischer2015Flownet:}, the flow estimation module in our framework predicts a motion field tailored to a specific task, \eg, frame interpolation or video denoising, as it is jointly trained with the corresponding video processing module.

Several papers have incorporated a learned motion estimation network in burst processing~\citep{tao2017detail,Liu2017Video}. In this paper, we move beyond to demonstrate not only how joint learning helps, but also why it helps. We show that a jointly trained network learns task-specific features for better video processing. For example, in video denoising, our \model learns to reduce the noise in the input, while traditional optical flow keeps the noisy pixels in the registered frame. \model also reduces artifacts near occlusion boundaries. Our goal in this paper is to build a standard framework for better understanding when and how task-oriented flow works.

To evaluate the proposed \toflow, we have also built a large-scale, high-quality video dataset for video processing. Most existing large video datasets, such as Youtube-8M~\citep{Abu-El-Haija2016Youtube}, are designed for high-level vision tasks like event classification. The videos are often of low resolutions with significant motion blurs, 
making them less useful for video processing. We introduce a new dataset, \data, for a systematic evaluation of video processing algorithms. \data consists of 89,800 
high-quality video clips (\ie 720p or higher) downloaded from Vimeo. We build three benchmarks from these videos for interpolation, denoising or deblocking, and super-resolution, respectively. We hope these benchmarks will also help improve learning-based video processing techniques with their high-quality videos and diverse examples.

This paper makes three contributions. First, we propose \model, a flow representation tailored to specific video processing tasks, significantly outperforming standard optical flow. 
Second, we propose an end-to-end trainable video processing framework that handles frame interpolation, video denoising, and video super-resolution. The flow network in our framework is fine-tuned by minimizing a task-specific, self-supervised loss.
Third, we build a large-scale, high-quality video processing dataset, \data.

\section{Related Work}
\label{sec:related}

\myparagraph{Optical flow estimation. }
Dated back to \cite{horn1981determining}, most optical flow algorithms have sought to minimize hand-crafted energy terms for image alignment and flow smoothness~\citep{memin1998dense,brox2004high,Brox2009Large,wedel2009structure}.
Current state-of-the-art methods like EpicFlow~\citep{Revaud2015Epicflow} or DC Flow~\citep{Xu2017Accurate} further exploit image boundary and segment cues to improve the flow interpolation among sparse matches.
Recently, end-to-end deep learning methods were proposed for 
faster inference~\citep{Fischer2015Flownet:,Ranjan2017Optical,yu2016back}. We use the same network structure for motion estimation as SpyNet~\citep{Ranjan2017Optical}. But instead of training it to minimize the flow estimation error, as SpyNet does, we train it jointly with a video processing network to learn a flow representation that is the best for a specific task.

\myparagraph{Low-level video processing.}
We focus on three video processing tasks: frame interpolation, video denoising, and video super-resolution.
Most existing algorithms in these areas explicitly estimate the dense correspondence among input frames, and then reconstruct the reference frame according to image formation models
for frame interpolation~\citep{baker2011database,werlberger2011optical,yu2013multi,jiang2018depth,sajjadi2018frame}, video super-resolution~\citep{liu2014bayesian,liao2015video}, and denoising~\citep{liu2010high,varghese2010video,maggioni2012video,mildenhall2018burst,godard2017deep}.
We refer readers to survey articles~\citep{nasrollahi2014super,ghoniem2010nonlocal} for comprehensive literature reviews on these flourishing research topics.

\myparagraph{Deep learning for video enhancement.}
Inspired by the success of deep learning, researchers have directly modeled enhancement tasks as regression problems without representing motions, and have designed deep networks for frame interpolation~\citep{Mathieu2016Deep,niklaus2017video,jiang2017super,niklaus2018context}, super-resolution~\citep{huang2015bidirectional,kappeler2016video,tao2017detail,bulat2018learn,ahn2018fast,jo2018deep}, denoising~\cite{mildenhall2018burst}, deblurring~\cite{yang2018multi,aittala2018burst}, rain drops removal~\citep{li2018video}, and video compression artifacts removal~\citep{lu2018deep}.

Recently, with differentiable image sampling layers in deep learning~\citep{Jaderberg2015Spatial}, motion information can be incorporated into networks and trained jointly. Such approaches have been applied to video interpolation~\citep{Liu2017Video}, light-field interpolation~\citep{wang2017light}, novel view synthesis~\citep{zhou2016view}, eye gaze manipulation \citep{Ganin2016DeepWarp:}, object detection~\citep{zhu2017flow}, denoising~\citep{wen2017joint}, and super-resolution~\citep{Caballero2017Real,tao2017detail,makansi2017end}. Although many of these algorithms also jointly train the flow estimation with the rest parts of network, there is no systematical study on the advantage of joint training. In this paper, we illustrate the advantage of the trained task-oriented flow through toy examples, and also demonstrate its superiority over general flow algorithm on various real-world tasks. We also present a general framework that can easily adapt to different video processing tasks. 

\section{Tasks}
\label{sec:task}

In the paper, we explore three video enhancement tasks: frame interpolation, video denoising/deblocking, and video super-resolution.

\myparagraph{Temporal frame interpolation. } Given a low frame rate video, a temporal frame interpolation algorithm generates a high frame rate video by synthesizing additional frames between two temporally neighboring frames. Specifically, let $I_1$ and $I_3$ be two consecutive frames in an input video, the task is to estimate the missing middle frame $I_2$. Temporal frame interpolation doubles the video frame rate, and can be recursively applied to generate even higher frame rates. 

\myparagraph{Video denoising/deblocking. } Given a degraded video with artifacts from either the sensor or compression, video denoising/deblocking aims to remove the noise or compression artifacts to recover the original video. This is typically done by aggregating information from neighboring frames. Specifically, Let $\{I_1, I_2, \dots, I_N\}$ be $N$ consecutive, degraded frames in an input video, the task of video denoising is to estimate the middle frame $I_\text{ref}^*$. For the ease of description, in the rest of paper, we simply call both tasks as video denoising. 

\myparagraph{Video super-resolution. } Similar to video denoising, given $N$ consecutive low-resolution frames as input, the task of video super-resolution is to recover the high-resolution middle frame. In this work, we first upsample all the input frames to the same resolution as the output using bicubic interpolation, and our algorithm only needs to recover the high-frequency component in the output image.

\section{Task-Oriented Flow for Video Processing}
\label{sect:model}

\begin{figure*}[t]
    \centering
    \includegraphics[width=\linewidth]{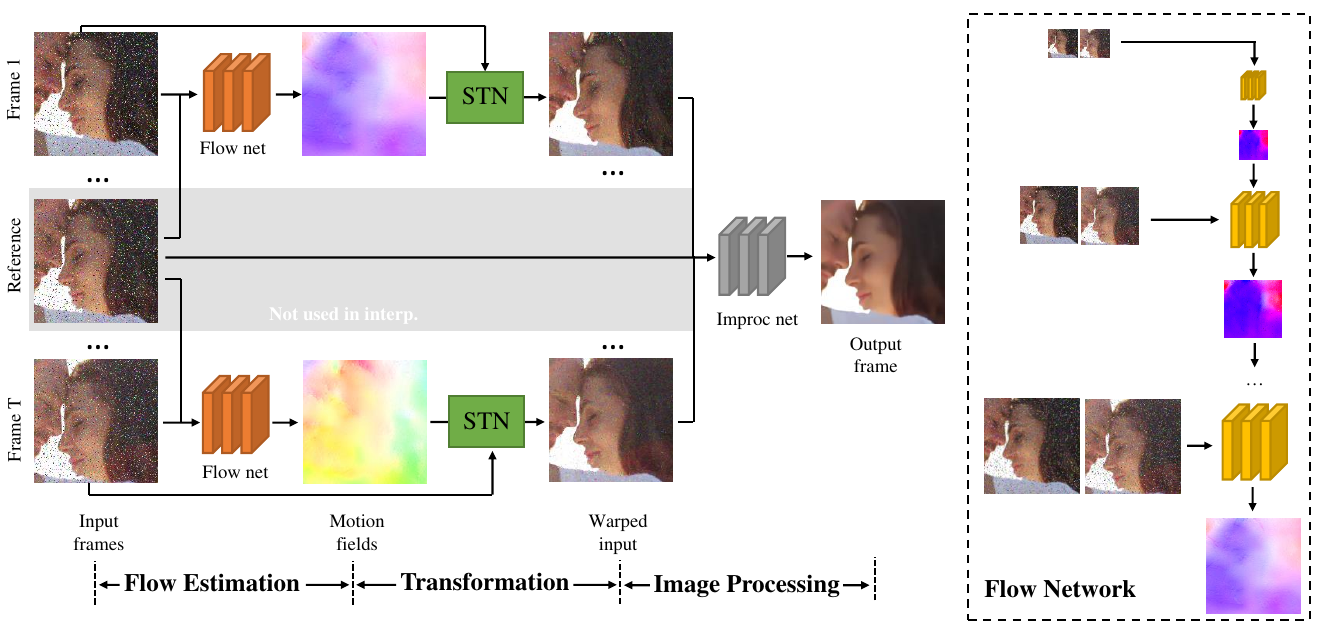}
    \caption{Left: our model using task-oriented flow for video processing. Given an input video, we first calculate the motion between frames through a task-oriented flow estimation network. We then warp input frames to the reference using spatial transformer networks, and aggregate the warped frames to generate a high-quality output image. Right: the detailed structure of flow estimation network (the orange network on the left).
    }
    \label{fig:pipeline}
\end{figure*}

Most motion-based video processing algorithms has two steps: motion estimation and image processing.
For example, in temporal frame interpolation, most algorithms first estimate how pixels move between input frames (frame 1 and 3), and then move pixels to the estimated location in the output frame (frame 2)~\citep{baker2011database}. Similarly, in video denoising, algorithms first register different frames based on estimated motion fields between them, and then remove noises by aggregating information from registered frames.

In this paper, we propose to use task-oriented flow (TOFlow) to integrate the two steps, which greatly improves the performance. To learn task-oriented flow, we design an end-to-end trainable network with three parts (\fig{fig:pipeline}): a flow estimation module that estimates the movement of pixels between input frames; an image transformation module that warps all the frames to a reference frame; and a task-specific image processing module that performs video interpolation, denoising, or super-resolution on registered frames. Because the flow estimation module is jointly trained with the rest of the network, it learns to predict a flow field that fits to a particular task.

\subsection{Toy Example}
\label{sec:toy_exmaple}

\begin{figure}[t]
    \centering
    \includegraphics[width=\columnwidth]{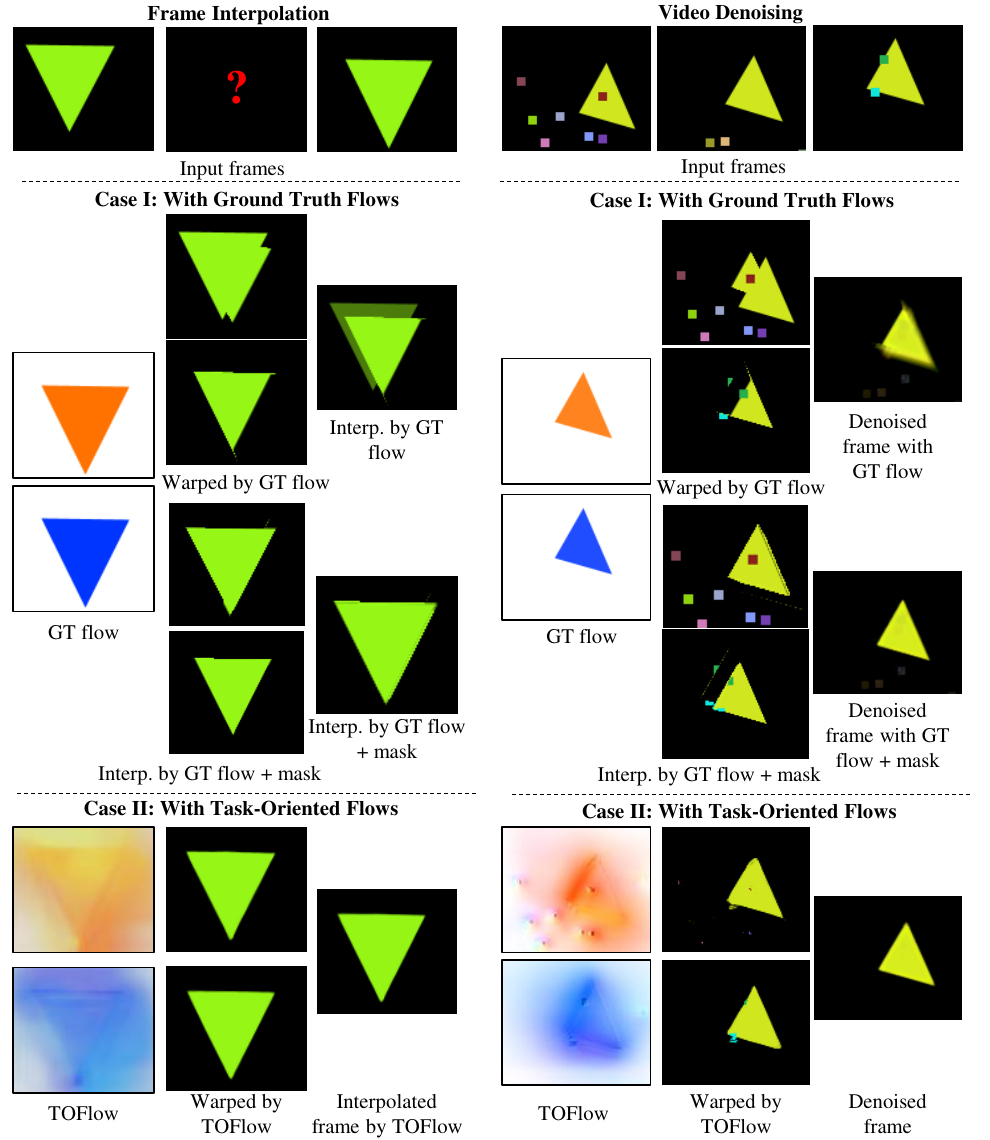}
    \caption{A toy example that demonstrates the effectiveness of task oriented flow over the traditional optical flow. See \sect{sec:toy_exmaple} for details. 
    }
    \label{fig:toy_example}
\end{figure}

Before discussing the details of network structure, we first start with two synthetic sequences to demonstrate why our \toflow can outperform traditional optical flows. The left of \fig{fig:toy_example} shows an example of frame interpolation, where a green triangle is moving to the bottom in front of a black background. If we warp both the first and the third frames to the second, even using the ground truth flow (Case I, left column), there is an obvious doubling artifact in the warped frames due to occlusion (Case I, middle column, top two rows), which is a well-known problem in the optical flow literature~\citep{baker2011database}. The final interpolation result based on these two warp frames still contains the doubling artifact (Case I, right column, top row). In contrast, \toflow does not stick to object motion: the background should be static, but it has non-zero motion (Case II, left column). With \toflow, however, there is barely any artifact in the warped frames (Case II, middle column) and the interpolated frame looks clean (Case II, right column). This is because \toflow not only synthesize the movement of visible object, but also guide how to inpaint occluded background region by copying pixels from its neighborhood. Also, if the ground truth occlusion mask is available, the interpolation result using ground truth flow will also contain little doubling artifacts (Case I, bottom rows). However, calculating the ground occlusion mask is even harder task than estimate flow, as it also requires inferring the correct depth ordering. On the other side, \toflow can handle occlusion and synthesize frames better than the ground truth flow without using ground truth occlusion masks and depth ordering information. 

Similarly, on the right of \fig{fig:toy_example}, we show an example of video denoising. The random small boxes in the input frames are synthetic noises. If we warp the first and the third frames to the second using the ground truth flow, the noisy patterns (random squares) remain, and the denoised frame still contains some noise (Case I, right column. There are some shadows of boxes on the bottom). But if we warp these two frames using \toflow (Case II, left column), those noisy patterns are also reduced or eliminated (Case II, middle column), and the final denoised frame base on them contains almost no noise, even better than the result by denoising results with ground truth flow and occlusion mask (Case I, bottom rows).  This also shows that \toflow learns to reduce the noise in input frames by inpainting them with neighboring pixels, which traditional flow cannot do.

Now we discuss the details of each module as follows. 

\subsection{Flow Estimation Module}
\label{sect:flow_net}

The flow estimation module calculates the motion fields between input frames. For a sequence with $N$ frames ($N=3$ for interpolation and $N=7$ for denoising and super-resolution), we select the middle frame as the reference. The flow estimation module consists of $N-1$ flow networks, all of which have the same structure and share the same set of parameters. Each flow network (the orange network in Figure~\ref{fig:pipeline}) takes one frame from the sequence and the reference frame as input, and predicts the motion between them.

We use the multi-scale motion estimation framework proposed by~\cite{Ranjan2017Optical} to handle the large displacement between frames. The network structure is shown in the right of Figure~\ref{fig:pipeline}. The input to the network are Gaussian pyramids of both the reference frame and another frame rather than the reference. At each scale, a sub-network takes both frames at that scale and upsampled motion fields from previous prediction as input, and calculates a more accurate motion fields. We uses 4 sub-networks in a flow network, three of which are shown Figure~\ref{fig:pipeline} (the yellow networks). 

There is a small modification for frame interpolation, where the reference frame (frame 2) is not an input to the network, but what it should synthesize. To deal with that, the motion estimation module for interpolation consists of two flow networks, both taking both the first and third frames as input, and predict the motion fields from the second frame to the first and the third respectively. With these motion fields, the later modules of the network can transform the first and the third frames to the second frame for synthesis.

\subsection{Image Transformation Module}
\label{sect:stn}

Using the predicted motion fields in the previous step, the image transformation module registers all the input frames to the reference frame. We use the spatial transformer networks~\citep{Jaderberg2015Spatial} (STN) for registration, which is a differentiable bilinear interpolation layer that synthesizes the new frame after transformation. Each STN transforms one input frame to the reference viewpoint, and all $N-1$ STNs forms the image transformation module. One important property of this module is that it can back-propagate the gradients from the image processing module to the flow estimation module, so we can learn a flow representation that adapts to different video processing tasks.

\subsection{Image Processing Module}
\label{sect:improc_net}

We use another convolutional network as the image processing module to generate the final output. 
For each task, we use a slightly different architecture. Please refer to appendices for details.

\myparagraph{Occluded regions in warped frames. } As mentioned \sect{sec:toy_exmaple}, occlusion often results in doubling artifacts in the warped frames. 
A common way to solve this problem is to mask out occluded pixels in interpolation, for example, \cite{Liu2017Video} proposed to use an additional network that estimates the occlusion mask and only uses pixels are not occluded. 

\begin{figure}[t]
    \centering
    \includegraphics[width=\columnwidth]{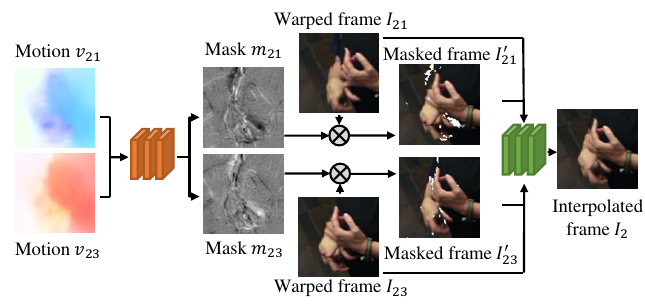}
    \caption{The structure of the mask network for interpolation}
    \label{fig:mask_net}
\end{figure}

Similar to~\cite{Liu2017Video}, we also tried the mask prediction network. It takes the two estimated motion fields as input, one from frame 2 to frame 1, and the other from frame 2 to frame 3 ($v_{21}$ and $v_{23}$ in \fig{fig:mask_net}). It predicts two occlusion masks: $m_{21}$ is the mask of the warped frame 2 from frame 1 ($I_{21}$), and $m_{23}$ is the mask of the warped frame 2 from frame 3 ($I_{23}$). The invalid regions in the warped frames ($I_{21}$ and $I_{23}$) are masked out by multiplying them with their corresponding masks. The middle frame is then calculated through another convolutional neural network with both the warped frames ($I_{21}$ and $I_{23}$) and the masked warped frames ($I_{21}'$ and $I_{23}'$) as input. Please refer to appendices for details.

\begin{figure}[t]
    \centering
    \includegraphics[width=\columnwidth]{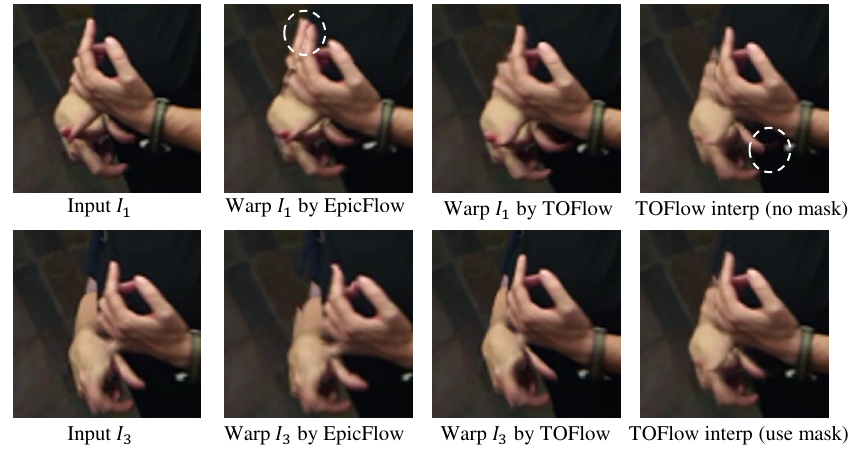}
    \caption{Comparison between Epicflow~\citep{Revaud2015Epicflow} and TOFlow interpolation (both with and without mask).}
    \label{fig:mask_example}
\end{figure}

An interesting observation is that, even without the mask prediction network, our flow estimation is mostly robust to occlusion. As shown in the third column of \fig{fig:mask_example}, the warped frames using \toflow has little doubling artifacts. Therefore, just from two warped frames without the learned masks, the network synthesizes a decent middle frame (the top image of the right most column). The mask network is optional, as it only removes some tiny artifacts.

\subsection{Training}
\label{sect:training}

To accelerate the training procedure, we first pre-train some modules of the network and then fine-tune all of them together. Details are described below.

\myparagraph{Pre-training the flow estimation network. }
Pre-training the flow network consists of two steps. First, for all tasks, we pre-train the \flownet network on the Sintel dataset~\citep{Butler2012naturalistic}, a realistically rendered video dataset with ground truth optical flow. 

In the second step, for video denoising and super-resolution, we fine-tune it with noisy or blurry input frames to improve its robustness to these input. For video interpolation, we fine-tune it with frames $I_1$ and $I_3$ from video triplets as input, minimizing the $l_1$ difference between the estimated optical flow and the ground truth flow $v_{23}$ (or $v_{21}$). This enables the flow network to calculate the motion from the unknown frame $I_2$ to frame $I_3$ given only frames $I_1$ and $I_3$ as input.

Empirically we find that this two-step pre-training can improve the convergence speed. Also, because the main purpose of pre-training is to accelerate the convergence, we simply use the $l_1$ difference between estimated optical flow and the ground truth as the loss function, instead of end-point error in flow literature~\citep{Brox2009Large,Butler2012naturalistic}. The choice loss function in the pre-training stage has a minor impact on the final result.

\myparagraph{Pre-training the mask network. }
We also pre-train our occlusion mask estimation network for video interpolation as an optional component of \procnet network before joint training. Two occlusion masks ($m_{21}$ and $m_{23}$) are estimated together with the same network and only optical flow $v_{21}, v_{23}$ as input. The network is trained by minimizing the $l_1$ loss between the output masks and pre-computed occlusion masks.

\myparagraph{Joint training. } After pre-training, we train all the modules jointly by minimizing the $l_1$ loss between recovered frame and the ground truth, without any supervision on estimated flow fields. For optimization, we use ADAM~\citep{Kingma2015Adam:} with a weight decay of $10^{-4}$. 
We run 15 epochs with batch size 1 for all tasks. 
The learning rate for denoising/deblocking and super-resolution is $10^{-4}$, and the learning rate for interpolation is $3\times10^{-4}$. 

\section{The \data Dataset}

\begin{figure*}[t]
    \centering
    \includegraphics[width=\linewidth]{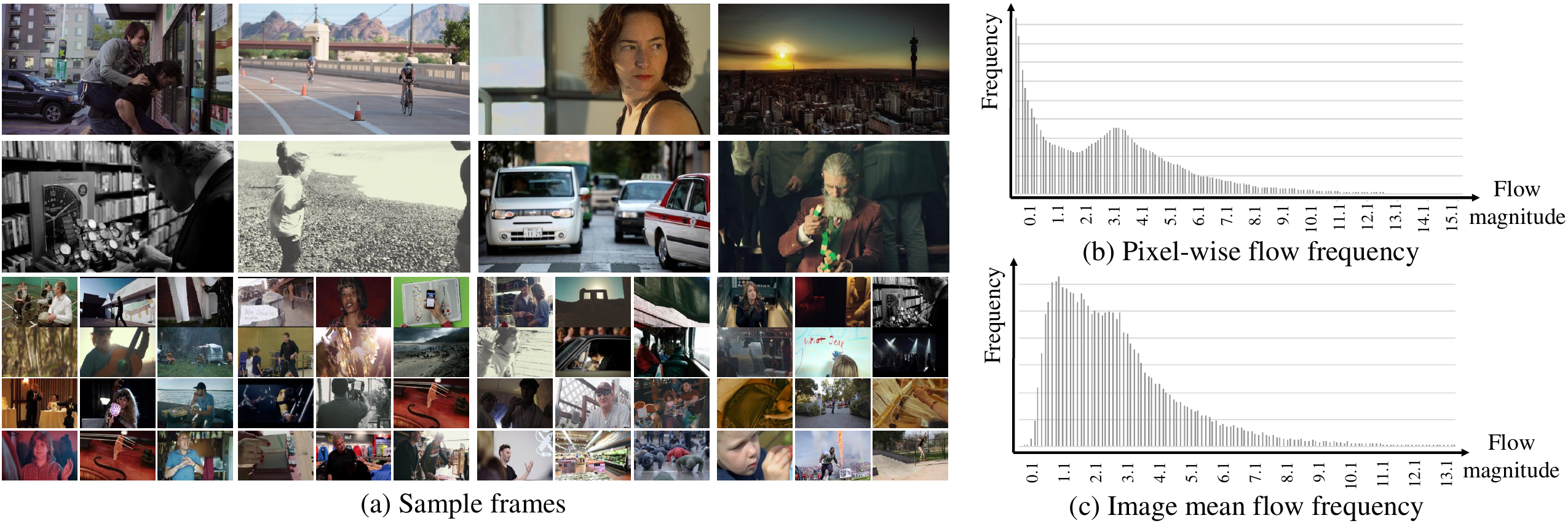}
    \caption{The \data dataset. (a) Sampled frames from the dataset, which show the high quality and wide coverage of our dataset; (b) The histogram of flow magnitude of all pixels in the dataset; (c) The histogram of mean flow magnitude of all images (the flow magnitude of an image is the average flow magnitude of all pixels in that image).}
    \label{fig:data}
\end{figure*}

To acquire high quality videos for video processing, previous methods~\citep{liu2014bayesian,liao2015video} took videos by themselves, resulting in video datasets that are small in size and limited in terms of content. Alternatively, we resort to Vimeo
where many videos are taken with professional cameras on diverse topics. In addition, we only search for videos without inter-frame compression (\eg, H.264), so that each frame is compressed independently, avoiding artificial signals introduced by video codecs. As many videos are composed of multiple shots, we use a simple threshold-based shot detection algorithm to break each video into consistent shots and further use GIST feature~\citep{Oliva2001Modeling} to remove shots with similar scene background.

As a result, we collect a new video dataset from Vimeo, consisting of 4,278 videos with 89,800 independent shots that are different from each other in content.
To standardize the input, we resize all frames to the fixed resolution 448$\times$256. As shown in \fig{fig:data}, frames sampled from the dataset contain diverse content for both indoor and outdoor scenes. 
We keep consecutive frames when the average motion magnitude is between 1--8 pixels. The right column of \fig{fig:data} shows the histogram of flow magnitude over the whole dataset, where the flow fields are calculated using SpyNet~\citep{Ranjan2017Optical}.

We further generate three benchmarks from the dataset for the three video enhancement tasks studied in this paper. 

\myparagraph{\interpbenchmark. }
We select 73,171 frame triplets from 14,777 video clips with the following three criteria for the interpolation task. First, more than 5\% pixels should have motion larger than 3 pixels between neighboring frames. This criterion removes static videos. 
Second, $l_1$ difference between the reference and the warped frame using optical flow (calculated using SpyNet) should be at most 15 intensity levels (the maximum intensity level of an image is 255). This removes frames with large intensity change, which are too hard for frame interpolation. Third, the average difference between motion fields of neighboring frames ($v_{21}$ and $v_{23}$) should be less than 1 pixel. This removes non-linear motion, as most interpolation algorithms, including ours, are based on linear motion assumption. 

\begin{figure*}[t]
    \centering
    \includegraphics[width=\linewidth]{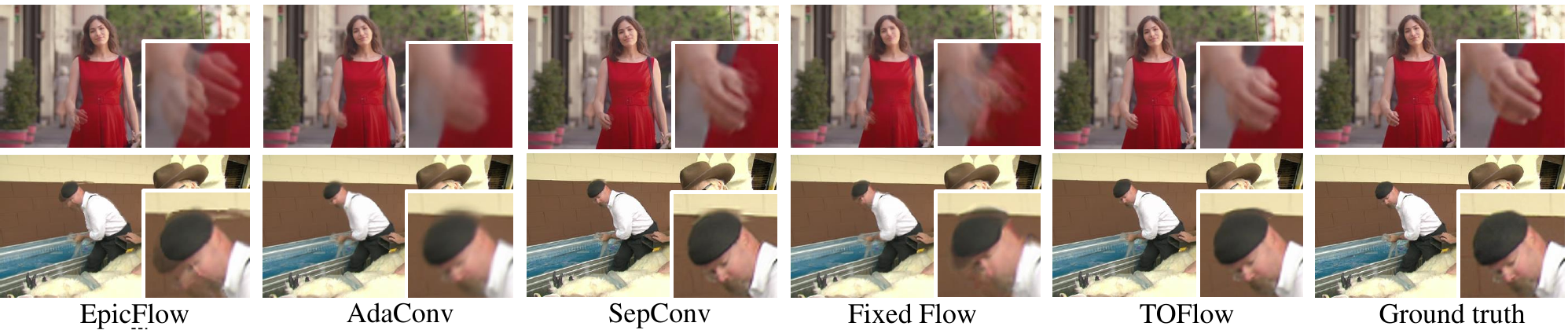}
    \caption{Qualitative results on frame interpolation. Zoomed-in views are shown in lower right.}
    \label{fig:interp}
\end{figure*}

\myparagraph{\denoisingbenchmark. }
We select 91,701 frame septuplets from 38,990 video clips for the denoising task, using the first two criteria introduced for the interpolation benchmark. 
For video denoising, we consider two types of noises: a Gaussian noise with a standard deviation of 0.1, and mixed noises including a 10$\%$ salt-and-pepper noise in addition to the Gaussian noise. For video deblocking, we compress the original sequences using FFmpeg with codec JPEG2000, format J2k, and quantization factor $q = \{20, 40, 60\}$.

\myparagraph{\srbenchmark. } We also use the same set of septuplets for denoising to build the \srbenchmark with down-sampling factor of 4: the resolution of input and output images are $112 \times 64$ and $448 \times 256$ respectively. To generate the low-resolution videos from high-resolution input, we use the MATLAB \textit{imresize} function, which first blurs the input frames using cubic filters and then downsamples videos using bicubic interpolation.

\section{Evaluation}
\label{sec:eval}

In this section, we evaluate two variations of the proposed network. The first one is to train each module separately: we first pre-train \flownet, and then train \procnet while fixing the flow module. This is similar to the two-step video processing algorithms, and we refer to it as \fflow. The other one is to jointly train all modules as described in~\sect{sect:training}, and we refer to it as \model. Both networks are trained on Vimeo benchmarks we collected. We evaluate these two variations on three different tasks and also compare with other state-of-the-art image processing algorithms.

\subsection{Frame Interpolation}
\label{sec:exp_interp}

\myparagraph{Datasets.}
We evaluate on three datasets: \interpbenchmark, the dataset used by~\cite{Liu2017Video} (DVF), and Middlebury flow dataset~\citep{baker2011database}.

\myparagraph{Metrics. } We use two quantitative measure to evaluate the performance of interpolation algorithms: peak signal-to-noise ratio (PSNR) and structural similarity (SSIM) index.

\myparagraph{Baselines. }
We first compare our framework with two-step interpolation algorithms.
For the motion estimation, we use \epicflow~\citep{Revaud2015Epicflow} and SpyNet~\citep{Ranjan2017Optical}.
To handle occluded regions as mentioned in~\sect{sect:improc_net}, we calculate the occlusion mask for each frame using the algorithm proposed by \cite{Zitnick2004High} and only use non-occluded regions to interpolate the middle frame.
Further, we compare with state-of-the-art end-to-end models, Deep Voxel Flow (DVF)~\citep{Liu2017Video}, Adaptive Convolution (AdaConv)~\citep{niklaus2017video}, and Separable Convolution (SepConv)~\citep{niklaus2017iccv}. At last, we also compare with \fflow, which is another baseline two-step interpolation algorithm\footnote{Note that \fflow or \toflow only uses 4-level structure of SpyNet for memory efficiency, while the original SpyNet network has 5 levels.}.

\begin{table}[t]
\centering
\setlength{\tabcolsep}{5pt}
\begin{tabular}{lcccc}
    \toprule
    \multirow{2}{*}{Methods} & \multicolumn{2}{c}{Vimeo Interp.} & \multicolumn{2}{c}{DVF Dataset}\\
    \cmidrule(lr){2-3}\cmidrule(lr){4-5}
    & PSNR & SSIM & PSNR & SSIM \\
    \midrule
    SpyNet   & 31.95 & 0.9601 & 33.60 & 0.9633\\
    EpicFlow & 32.02 & 0.9622
                           & 33.71 & 0.9635\\
    DVF         & 33.24 & 0.9627 
                           & 34.12 & 0.9631\\
    AdaConv       & 32.33 & 0.9568
                           & --- & --- \\
    SepConv       & 33.45 & 0.9674
                           & \textbf{34.69} & 0.9656 \\
    \midrule
    \fflow & 29.09 & 0.9229
                        & 31.61 & 0.9544\\
    \fflow + Mask & 30.10 & 0.9322
                             & 32.23 & 0.9575\\
    \midrule
    \model          & 33.53 & 0.9668
                    & 34.54 & 0.9666 \\
    \model + Mask & \textbf{33.73} & \textbf{0.9682}
                & 34.58 & \textbf{0.9667} \\
    \bottomrule
\end{tabular}
\caption{Quantitative results of different frame interpolation algorithms on the Vimeo interpolation test set and the DVF test set~\citep{Liu2017Video}.}
\label{tbl:interp}

\end{table}
\begin{figure*}[t]
    \centering
    \includegraphics[width=\linewidth]{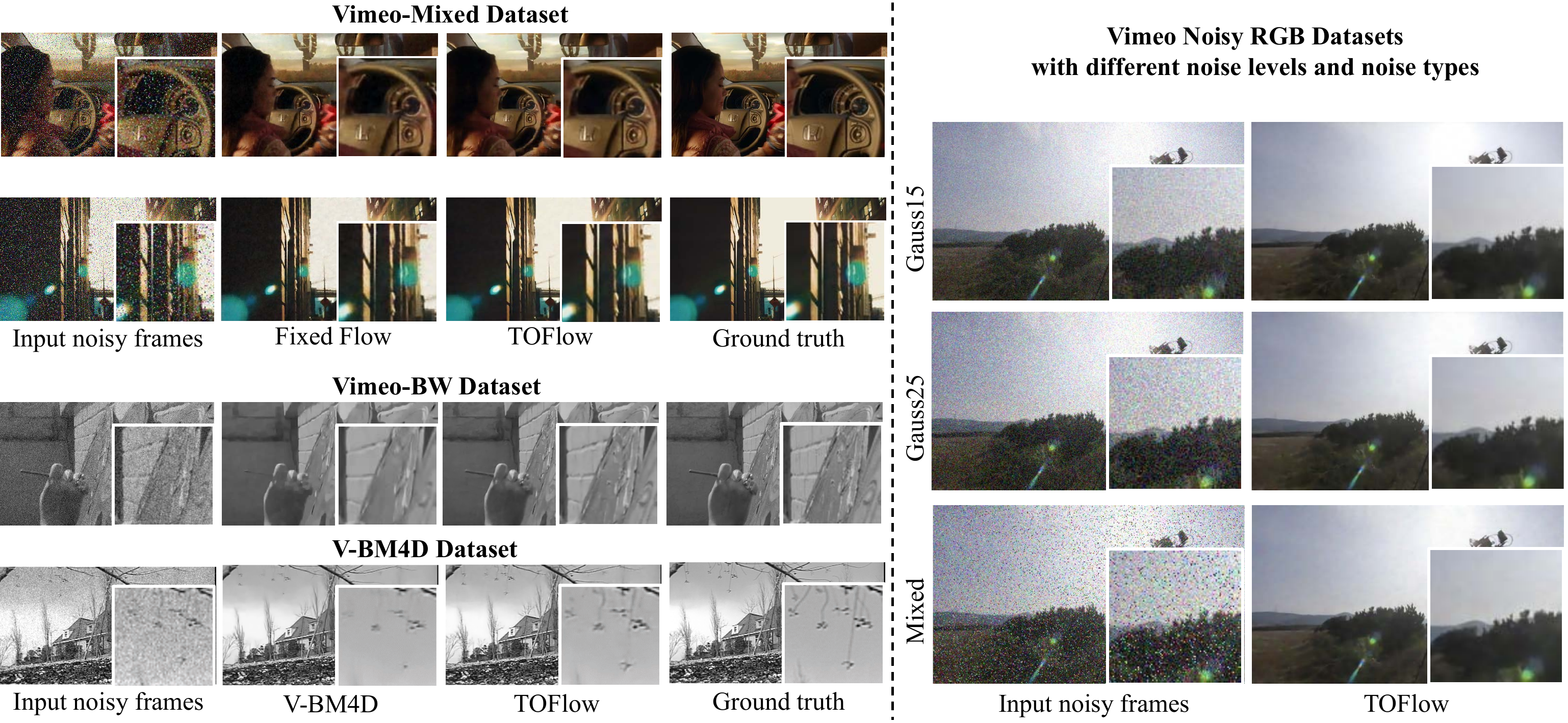}
    \caption{Qualitative results on video denoising. The differences are clearer when zoomed-in. 
    }
     \label{fig:denoise}
\end{figure*}

\myparagraph{Results. }
\tbl{tbl:interp} shows our quantitative results\footnote{We did not evaluate AdaConv on DVF dataset, as neither the implementation of AdaConv nor the DVF dataset is publicly available.}. On \interpbenchmark, \model in general outperforms the others interpolation algorithms, both the traditional two-step interpolation algorithms (\epicflow and SpyNet) and recent deep-learning based algorithms (DVF, AdaConv, and SepConv), with a significant margin. Though our model is trained on our \data dataset, it also outperforms DVF on DVF dataset in both PSNR and SSIM. There is also a significant boost over \fflow, showing that the network does learn a better flow representation for interpolation during joint training.

\fig{fig:interp} also shows qualitative results. All the two-step algorithms (\epicflow and \fflow) generate a doubling artifacts, like the hand in the first row or the head in the second row. AdaConv on the other sides does not have the doubling artifacts, but it tends to generate blurry output, by directly synthesizing interpolated frames without a motion module. SepConv increases the sharpness of output frame compared with AdaConv, but there are still artifacts (see the hat on the bottom row). Compared with these methods, \model correctly recovers sharper boundaries and fine details even in presence of large motion.  

\tbl{tbl:middlebury} shows a qualitative comparison of the proposed algorithms with the best four alternatives on Middlebury \citep{baker2011database}. We use the sum of square difference (SSD) reported on the official website as the evaluation metric. \model performs better than other interpolation networks. 

\begin{table}[t]
\centering
\small
\setlength{\tabcolsep}{3pt}
\begin{tabular}{lccccc}
    \toprule
    \multirow{2}{*}{Methods} & \multirow{2}{*}{PMMST} & \multirow{2}{*}{DeepFlow} & \multirow{2}{*}{SepConv} & \multirow{2}{*}{TOFlow} & TOFlow\\
    & & & & & Mask\\
    \midrule
    All  & 5.783 & 5.965 & 5.605 & 5.67 & \textbf{5.49}\\
    Discontinuous  & 9.545 & 9.785 & 8.741 & 8.82 & \textbf{8.54}\\
    Untextured & 2.101 & \textbf{2.045} & 2.334 & 2.20 & 2.17\\
    \bottomrule
\end{tabular}
\caption{Quantitative results of five frame interpolation algorithms on Middlebury flow dataset~\citep{baker2011database}: PMMST~\citep{xu2015pm}, SepConv~\citep{niklaus2017iccv}, DeepFlow~\citep{Liu2017Video}, and our TOFlow (with and without mask). Follow the convention of Middlebury flow dataset, we reported the square root error (SSD) between ground truth image and interpolated image in 1) entire images, 2) regions of motion discontinuities, and 3) regions without texture.}
\label{tbl:middlebury}

\end{table}

\begin{table*}
\begin{tabular}{lcccccc}
    \toprule
    \multirow{2}{*}{Methods} &
    \multicolumn{2}{c}{Vimeo-Gauss15} & 
    \multicolumn{2}{c}{Vimeo-Gauss25} & \multicolumn{2}{c}{Vimeo-Mixed} \\
    \cmidrule(lr){2-3}\cmidrule(lr){4-5}\cmidrule(lr){6-7}
    & PSNR & SSIM & PSNR & SSIM & PSNR & SSIM \\
    \midrule
    \fflow & 36.25 & 0.9626
               & 34.74 & 0.9411
               & 31.85 & 0.9089 \\
    \midrule
    \model         & \textbf{36.63} & \textbf{0.9628}
                   & \textbf{34.89} & \textbf{0.9518}
                   & \textbf{33.51} & \textbf{0.9395}
                   \\
    \bottomrule
\end{tabular}
\begin{tabular}{lcccc}
    \toprule
    \multirow{2}{*}{Methods} & 
    \multicolumn{2}{c}{Vimeo-BW} & \multicolumn{2}{c}{V-BM4D} \\
    \cmidrule(lr){2-3}\cmidrule(lr){4-5}
    & PSNR & SSIM & PSNR & SSIM \\
    \midrule
    \vbm4d 
                   & 33.17 & 0.8756
                   & \textbf{30.63} & 0.8759 \\
    \midrule
    \model  
                   & \textbf{36.75} & \textbf{0.9275}
                   & 30.36 & \textbf{0.8855}\\
    \bottomrule
\end{tabular}
\caption{Quantitative results on video denoising. Left: Vimeo RGB datasets with three different types of noise; Right: two grayscale dataset: Vimeo-BW and \vbm4d. 
}
\label{tbl:denoise}
\end{table*}

\subsection{Video Denoising/Deblocking}
\label{sec:denoise}

\myparagraph{Setup. }
We first train and evaluate our framework on Vimeo denoising benchmark, with three types of noises: Gaussian noise with standard deviation of 15 intensity levels (Vimeo-Gauss15), Gaussian noise with standard deviation of 25  (Vimeo-Gauss25), and mixture of Gaussian noise and 10\% salt-and-pepper noise  (Vimeo-Mixed). To compare our network with \vbm4d~\citep{maggioni2012video}, which is a monocular video denoising algorithm, we also transfer all videos in Vimeo Denoising Benchmark to grayscale to create Vimeo-BW (Gaussian noise only), and retrain our network on it. We also evaluate our framework on the a mono video dataset in \vbm4d.

\myparagraph{Baselines.}
We compare our framework with the \vbm4d, with the standard deviation of Gaussian noise as its additional input on two grayscale datasets (Vimeo-BW and \vbm4d). 
As before, we also compare with the \fflow variant of our framework on three RGB datasets (Vimeo-Gauss15, Vimeo-Gauss25, and Vimeo-Mixed). 

\begin{table}
\setlength{\tabcolsep}{4pt}
\small
\centering
\begin{tabular}{lC{0.85cm}C{0.9cm}C{0.85cm}C{0.9cm}C{0.85cm}C{0.9cm}}
    \toprule
    \multirow{2}{*}{Methods} & 
    \multicolumn{2}{c}{\makecell{Vimeo-Blocky \\ (q=20)}} &
    \multicolumn{2}{c}{\makecell{Vimeo-Blocky \\ (q=40)}} & 
    \multicolumn{2}{c}{\makecell{Vimeo-Blocky \\ (q=60)}}\\
    \cmidrule(lr){2-3}\cmidrule(lr){4-5}\cmidrule(lr){6-7}
    & PSNR & SSIM & PSNR & SSIM & PSNR & SSIM\\
    \midrule
    V-BM4D & 35.75 & 0.9587 & 33.72 & 0.9402 & 32.67 & 0.9287\\
    Fixed flow & 36.52 & 0.9636 & 34.50 & 0.9485 & 33.06 & 0.9168\\
    \midrule
    \model  & \textbf{36.92} & \textbf{0.9663} & \textbf{34.97} & \textbf{0.9527} & \textbf{34.02} & \textbf{0.9447} \\
    \bottomrule
\end{tabular}
\normalsize
\caption{Results on video deblocking. 
}
\label{tbl:deblock}
\end{table}

\begin{figure*}[t]
\centering
\includegraphics[width=\linewidth]{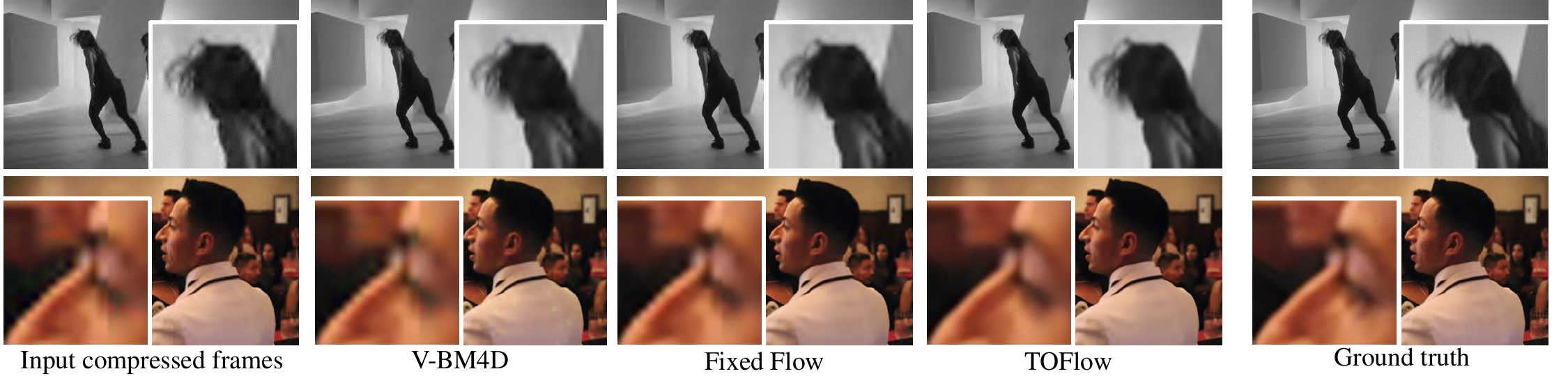}
\caption{Qualitative results on video deblocking. The differences are clearer when zoomed-in.}
\label{fig:deblock}
\end{figure*}

\begin{figure*}[t]
\centering
\includegraphics[width=\linewidth]{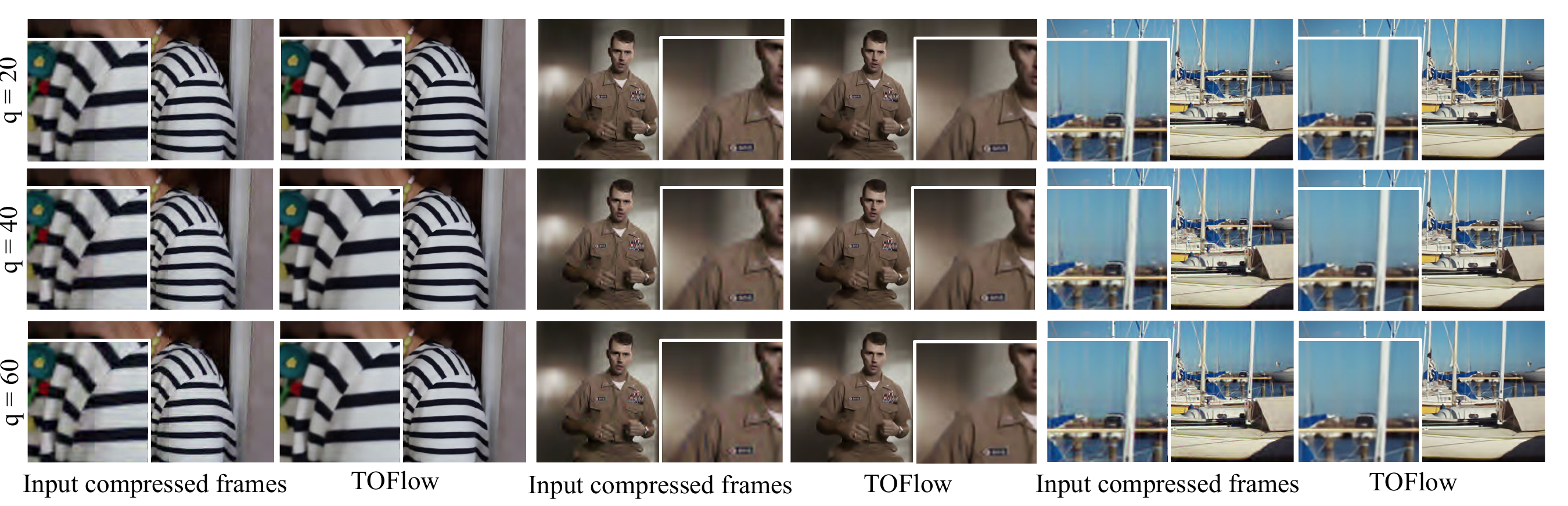}
\caption{Results on frames with different encoding qualities. The differences are clearer when zoomed-in.}
\label{fig:deblock_multiple_q}
\end{figure*}

\myparagraph{Results.}
We first evaluate \model on the Vimeo dataset with three different noise levels (\tbl{tbl:denoise}). \model outperforms \fflow by a significant margin, demonstrating the effectiveness of joint training. Also, when the noise level increases to 25 or when additional salt-and-pepper noise is added, the PSNR of \model is still round 34dB, showing its robustness to different noise levels. This is qualitatively demonstrated in the right half of \fig{fig:denoise}. 

On two grayscale datasets, Vimeo-BW and \vbm4d, \model outperforms \vbm4d in SSIM. Here we do not fine-tune it on \vbm4d. Though \model only achieves a comparable performance with \vbm4d in PSNR, the output of \model is much sharper than \vbm4d. As shown in \fig{fig:denoise}, the details of the beard and collar are kept in the denoised frame by \model (the mid left of \fig{fig:denoise}), and leaves on the tree are also clearer (the bottom left of \fig{fig:denoise}). Therefore, \model beats \vbm4d in SSIM, which better reflects human's perception than PSNR.

For video deblocking, \tbl{tbl:deblock} shows that \model outperforms \vbm4d. \fig{fig:deblock} also shows the qualitative comparison between \model, \fflow, and \vbm4d. Note that the compression artifacts around the girl's hair (top) and the man's nose (bottom) are completely removed by \model. The vertical line around the man's eye (bottom) due to a blocky compression is also removed by our algorithm. To demonstrate the robustness of our algorithms on video deblocking with different quantization levels, we also evaluate the three algorithms on input videos generated under three different quantization levels, and \model consistently outperforms other two baselines. \fig{fig:deblock_multiple_q} also shows that when the quantization level increases, the deblocking output remains mostly the same, suggesting the robustness of \model.

\begin{figure*}[t]
    \centering
    \includegraphics[width=\linewidth]{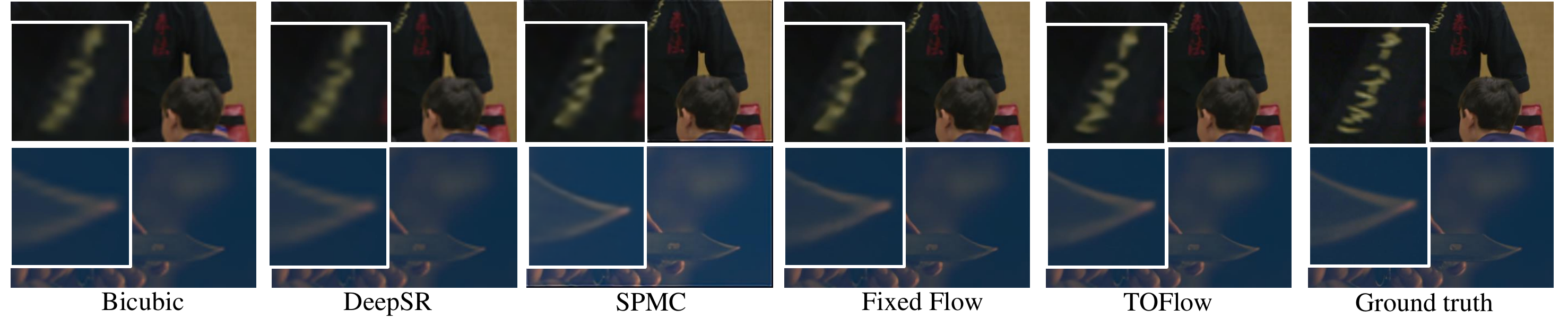}
    \caption{Qualitative results on super-resolution. Close-up views are shown on the top left of each result. The differences are clearer when zoomed-in.}
    \label{fig:sr}
\end{figure*}

\subsection{Video Super-Resolution}

\myparagraph{Datasets. }
We evaluate our algorithm on two dataset: \srbenchmark and the dataset provided by \cite{liu2011bayesian} (BayesSR). The later one consists of four sequences, each having 30 to 50 frames. \srbenchmark only contains 7 frames, so there is no full-clip evaluation for it.

\myparagraph{Baselines. }
We compare our framework with bicubic up-sampling, three 
video SR algorithms: BayesSR (we use the version provided by \cite{ma2015handling}), DeepSR~\citep{liao2015video}, and SPMC~\citep{tao2017detail}, as well as a baseline with a \fflow estimation module. Both BayesSR and DeepSR can take various number of frames as input. Therefore, on BayesSR dataset, we report two numbers: one on the whole sequence, the other on the seven frames in the middle, as SPMC, \model, and \fflow only take 7 frames as input.

\begin{table}[t]
\centering
\setlength{\tabcolsep}{4pt}
\begin{tabular}{llcccc}
    \toprule
    \multirow{2}{*}{Input} & \multirow{2}{*}{Methods} & \multicolumn{2}{c}{Vimeo-SR} & \multicolumn{2}{c}{BayesSR} \\
    \cmidrule(lr){3-4}\cmidrule(lr){5-6}
    &  & PSNR & SSIM & PSNR & SSIM \\
    \midrule
    \multirow{2}{*}{Full Clip} 
    & DeepSR
    & --- & --- 
    & 22.69 & 0.7746\\
    & BayesSR
    & --- & --- 
    & \textbf{24.32} & \textbf{0.8486}\\
    \midrule
    1 Frame 
    & Bicubic                
    & 29.79 & 0.9036
    & 22.02 & 0.7203 \\
    \midrule
    \multirow{5}{*}
    {7 Frames}
    & DeepSR
    & 25.55 & 0.8498
    & 21.85 & 0.7535\\
    & BayesSR
    & 24.64 & 0.8205
    & 21.95 & 0.7369 \\
    & SMPC
    & 32.70 & 0.9380
    & 21.84 & 0.7990 \\
    & \fflow
    & 31.81 & 0.9288
    & 22.85 & 0.7655 \\
    \cmidrule{2-6}
    & \model
    & \textbf{33.08} & \textbf{0.9417}
    & \textbf{23.54} & \textbf{0.8070} \\
    \bottomrule
\end{tabular}
\caption{Results on video super-resolution. Each clip in Vimeo-SR contains 7 frames, and each clip in BayesSR contains 30--50 frames. 
}
\label{tbl:sr}
\end{table}
\begin{table}
\centering
\begin{tabular}{cccccc}
    \toprule
    \multicolumn{2}{c}{3 Frames} &
    \multicolumn{2}{c}{5 Frames} & 
    \multicolumn{2}{c}{7 Frames}\\
    \cmidrule(lr){1-2}\cmidrule(lr){3-4}\cmidrule(lr){5-6}
    PSNR & SSIM & PSNR & SSIM & PSNR & SSIM\\
    \midrule
    32.66 & 0.9375 & 33.04 & 0.9415 & \textbf{33.08} & \textbf{0.9417} \\
    \bottomrule
\end{tabular}
\caption{Results on video super-resolution with a different number of input frames.}
\label{tbl:sr_frames}
\end{table}\textbf{}

\begin{table}[t]
\centering
\begin{tabular}{cccccc}
    \toprule
    \multicolumn{2}{c}{Cubic Kernel} &\multicolumn{2}{c}{Box Kernel} & \multicolumn{2}{c}{Gaussian Kernel} \\
    \cmidrule(lr){1-2}\cmidrule(lr){3-4}\cmidrule(lr){5-6}
    PSNR & SSIM & PSNR & SSIM & PSNR & SSIM \\
    \midrule
    33.08 & 0.9417 & 32.08 & 0.9372 & 31.15 & 0.9314 \\
    \bottomrule
\end{tabular}
\caption{Results of \toflow on video super-resolution when different downsampling kernels are used for building the dataset. 
}
\label{tbl:sr_kernels}
\end{table}

\myparagraph{Results. }
\tbl{tbl:sr} shows our quantitative results. Our algorithm performs better than baseline algorithms when using 7 frames as input, and it also achieves comparable performance to BayesSR when BayesSR uses all 30--50 frames as input while our framework only uses 7 frames. 
We show qualitative results in \fig{fig:sr}. Compared with either DeepSR or \fflow, the jointly trained \toflow generates sharper output. Notice the text on the cloth (top) and the tip of the knife (bottom) are clearer in the high-resolution frame synthesized by \model.
This shows the effectiveness of joint training.

To better understand how many input frames are sufficient for super-resolution, we also train our \toflow with different number of input frames, as shown in \tbl{tbl:sr_frames}. There is a big improvement when switching from 3-frame to 5-frame, and the improvement becomes minor when further switching to 7-frame. Therefore, 5 or 7 frames should be enough for super-resolution.

Besides, the down-sampling kernels (a.k.a. the point-spread function) used to create the low-resolution images may also affect the performance super-resolution~\citep{liao2015video}. To evaluate how down-sampling kernels affect the performance of our algorithm, we evaluate on three different kernels: cubic kernels, box down-sampling kernels, and Gaussian kernels with a variance of 2 pixels), and \tbl{tbl:sr_kernels} shows the result. There is a 1 dB drop in PSNR when switching to box kernels, and another 1 dB drop when switching to Gaussian kernels. This is because that down-sampling kernels remove high-frequency aliasing in low-resolution input images, making super-resolution harder. In most of experiments here, we follow the convention in previous multi-frame super-solution papers~\citep{liu2011bayesian,tao2017detail}, which creates low-resolution images through bicubic interpolation with no blur kernels. However, the results with blur kernels are also interesting, as it is closer the actual formation of low-resolution images captured by cameras.

\begin{figure}[t]
\centering
\includegraphics[width=\linewidth]{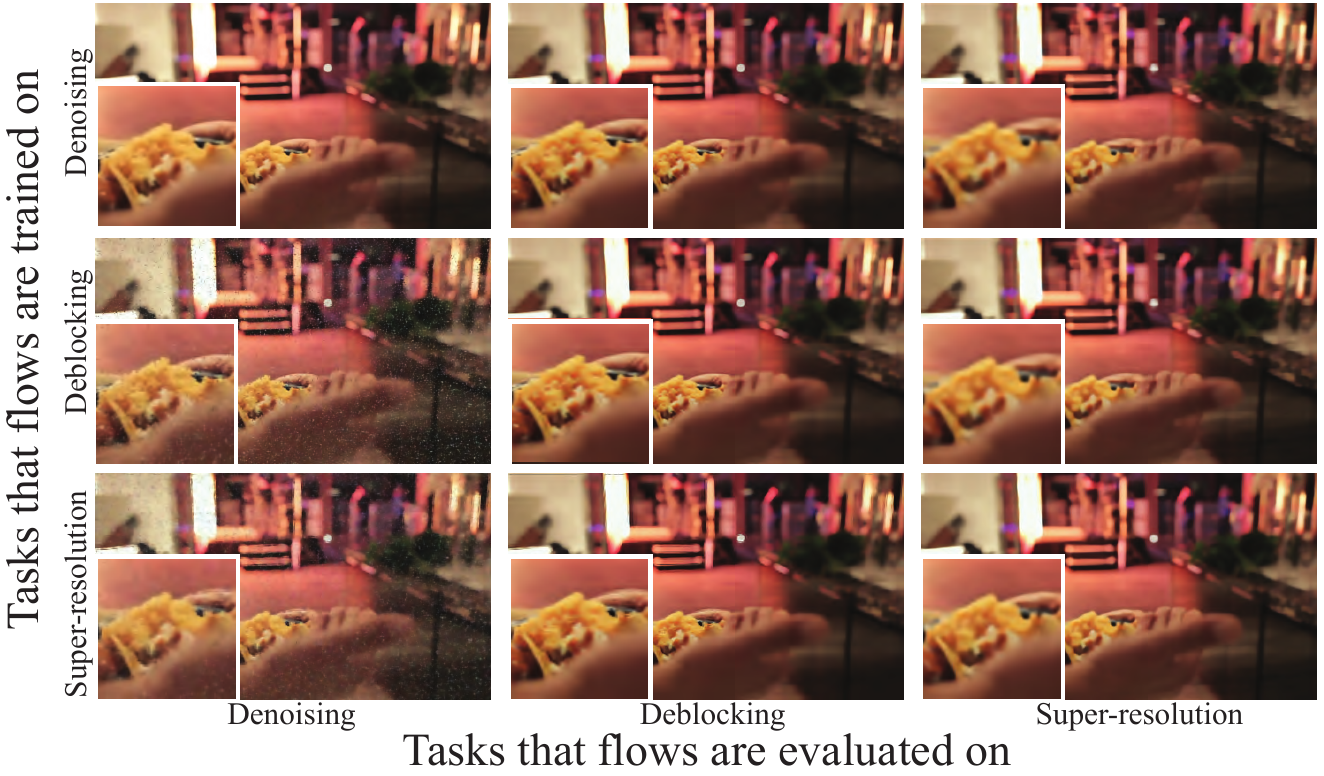}
\caption{Qualitative results of \toflow on tasks including but not limited to the one it was trained on.}
\label{fig:switch_flow}
\end{figure}

In all the experiments, we train and evaluate our network on an NVIDIA Titan X GPU. For an input clip with resolution $256 \times 448$, our network takes about 200ms for interpolation and 400ms for denoising or super-resolution (the resolution of the input to the super-resolution network is $64 \times 112$), where the flow module takes 18 ms for each estimated motion field.

\begin{figure*}[t]
    \centering
    \includegraphics[width=\linewidth]{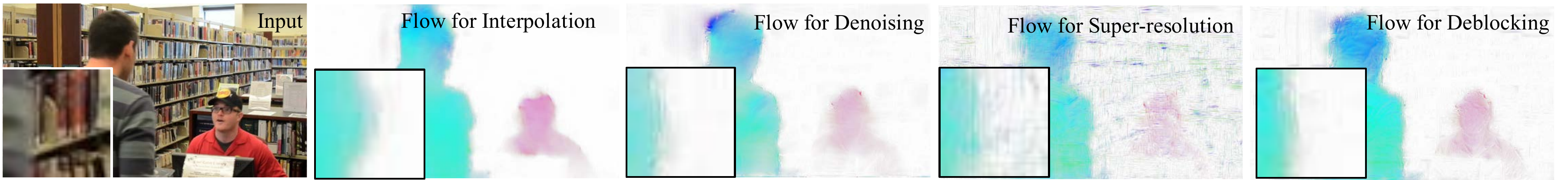}
    \caption{Visualization of motion fields for different tasks.}
     \label{fig:flowvis}
\end{figure*}

\subsection{Flows Learned from Different Tasks}

We now compare and contrast the flow learned from different tasks to understand if learning flow in such a task-oriented fashion is necessary.

We conduct an ablation study by replacing the flow estimation network in our model by a flow network trained on a different task (\fig{fig:switch_flow} and \tbl{tbl:switch_flow}). There is a significant performance drop when we use a flow network that is not trained on that task. For example, with the flow network trained on deblocking or super-resolution, the performance of the denoising algorithm drops by 5dB, and there are noticeable noises in the images (the first column of \fig{fig:switch_flow}). There are also ringing artifacts when we apply the flow network trained on super-resolution for deblocking (\fig{fig:switch_flow} row 2, col 2). Therefore, our task-oriented flow network is indeed tailored to a specific task. Besides, in all these three tasks, \fflow performs better than \model if trained and tested on different tasks, but worse than \model if trained and tested on the same task. This suggests that joint training improves the performance of a flow network on one task, but decreases its performance on the others. 

\fig{fig:flowvis} contrasts the motion fields learned from different tasks: the flow field for interpolation is very smooth, even on the occlusion boundaries, while the flow field for super-resolution has artificial movements along the texture edges. This indicates that the network may learn to encode different information that is useful for different tasks in the learned motion fields.

\begin{table}[t]
\centering
\small
\begin{tabular}{lccc}
    \toprule
    \multirow{2}{*}{Tasks trained on} & \multicolumn{3}{c}{Tasks evaluated on} \\
    \cmidrule{2-4}
    & Denoising & Deblocking & Super-resolution \\
    \midrule
    Denoising & \textbf{34.89} & 36.13 & 31.30 \\
    Deblocking & 25.74 & \textbf{36.92} & 31.86 \\
    Super-resolution & 25.99 & 31.86 & \textbf{33.08} \\
    \midrule
    \fflow & 34.74 & 36.52 & 31.81 \\
    \epicflow & 30.43 & 30.09 & 28.05 \\
    \bottomrule
\normalsize
\end{tabular}
\vspace{-7pt}
\caption{PSNR of \toflow on tasks including but not limited to the one it was trained on.}
\label{tbl:switch_flow}
\end{table}

\subsection{Accuracy of Retrained Flow}

As shown in the \fig{fig:teaser}, tailoring a motion estimation network to a specific task will reduce the accuracy of the estimated flow. To verify that, we evaluate the flow estimation accuracy on the Sintel Flow Dataset by~\cite{Butler2012naturalistic}. Three variants of flow estimation networks are tested: first, a network pre-trained on the Flying Chair dataset, second, the network after fine-tuning on denoising, and third, the network after fine-tuning on super-resolution. All fine-tuning is on the \data dataset. As shown in \tbl{tbl:sintel}, the accuracy of \toflow is much worse than either \epicflow~\citep{Revaud2015Epicflow} or \fflow\footnote{The EPE of \fflow on Sintel dataset is different from EPE of SpyNet~\citep{Ranjan2017Optical} reported on Sintel website, as it is trained differently from SpyNet as we mentioned before.}, but as shown in \tbl{tbl:denoise}, \ref{tbl:deblock}, and \ref{tbl:sr}, \toflow outperforms \fflow on specific tasks. This is consistent with the intuition that \toflow is a motion representation that does not match the actual object movement, but leads to better video processing results.

\begin{table}[t]
\centering
\small
\vspace{-4pt}\begin{tabular}{lcccc}
    \toprule
    \multirow{2}{*}{Methods} & \toflow & \toflow & Fixed & Epic\\
    & Denoising & Super-resolution & Flow & Flow \\
    \midrule
    All & 16.638 & 16.586 & 14.120 & \textbf{4.115} \\
    Matched & 11.961 & 11.851 & 9.322 & \textbf{1.360}	\\
    Unmatched & 54.724 & 55.137 & 53.117 & \textbf{26.595} \\
    \bottomrule
\end{tabular}
\caption{End-point-error (EPE) of estimated flow fields on the Sintel dataset. We evaluate \toflow (trained on two different tasks), \fflow, and \epicflow~\citep{Revaud2015Epicflow}. We report errors over full images, matched regions, and unmatched regions.}
\label{tbl:sintel}
\end{table}

\subsection{Different Flow Estimation Network Structure}

Our task-oriented video processing pipeline is not limited to one flow estimation network structure, although in all previous experiments, we use SpyNet by~\cite{Ranjan2017Optical} as the flow estimation module for its memory efficiency. To demonstrate the generalization ability of our framework, we also experiment with the FlowNetC~\citep{Fischer2015Flownet:} structure, and evaluate it on video denoising, deblocking, and super-resolution. Because FlowNetC has larger memory consumption, we only estimate flow at 256$\times$192 and upsample it to the target resolution. Its performance is therefore worse than the model using SpyNet, as shown in \tbl{tbl:flownet}. Still, in all these three tasks, \toflow outperforms than \fflow. This demonstrates the generalization ability of the \toflow framework to other flow estimation modules.

\begin{table}
\setlength{\tabcolsep}{0.12cm}
\footnotesize
\centering
\begin{tabular}{lC{0.85cm}C{0.9cm}C{0.85cm}C{0.9cm}C{0.85cm}C{0.9cm}}
    \toprule
    \multirow{2}{*}{Methods} & 
    \multicolumn{2}{c}{Denoising} &
    \multicolumn{2}{c}{Deblocking} & 
    \multicolumn{2}{c}{Super-resolution}\\
    \cmidrule(lr){2-3}\cmidrule(lr){4-5}\cmidrule(lr){6-7}
    & PSNR & SSIM & PSNR & SSIM & PSNR & SSIM\\
    \midrule
    \fflow & 24.685 & 0.8297 & 36.028 & 0.9672 & 31.834 & 0.9291 \\
    \model & \textbf{24.689} & \textbf{0.8374} & \textbf{36.496} & \textbf{0.9700} & \textbf{33.010} & \textbf{0.9411} \\
    \bottomrule
\end{tabular}
\normalsize
\caption{Results of \model on three different tasks, using FlowNetC~\citep{Fischer2015Flownet:} as the motion estimation module.}
\label{tbl:flownet}
\end{table}

\section{Conclusion}
\label{sec:dis}

In this work, we have proposed a novel video processing model that exploits task-oriented motion cues. Traditional video processing algorithms normally consist of two steps: motion estimation and video processing based on estimated motion fields. However, a genetic motion for all tasks might be sub-optimal and the accurate motion estimation would be neither necessary nor sufficient for these tasks. Our self-supervised, task-oriented flow (\toflow) bypasses this difficulty by modeling motion signals in the loop. 
To evaluate our algorithm, we have also created a new dataset, \data, for video processing. Extensive experiments on temporal frame interpolation, video denoising/deblocking, and video super-resolution demonstrate the effectiveness of \toflow.

\myparagraph{Acknowledgements. }
This work is supported by NSF RI \#1212849, NSF BIGDATA \#1447476, Facebook, Shell Research, and Toyota Research Institute. This work was done when Tianfan Xue and Donglai Wei were graduate students at MIT CSAIL.

\bibliographystyle{spbasic}      
\bibliography{toflow}   

\newpage
\section*{Appendices}

\begin{figure*}[p]
    \centering
    \includegraphics[width=\linewidth]{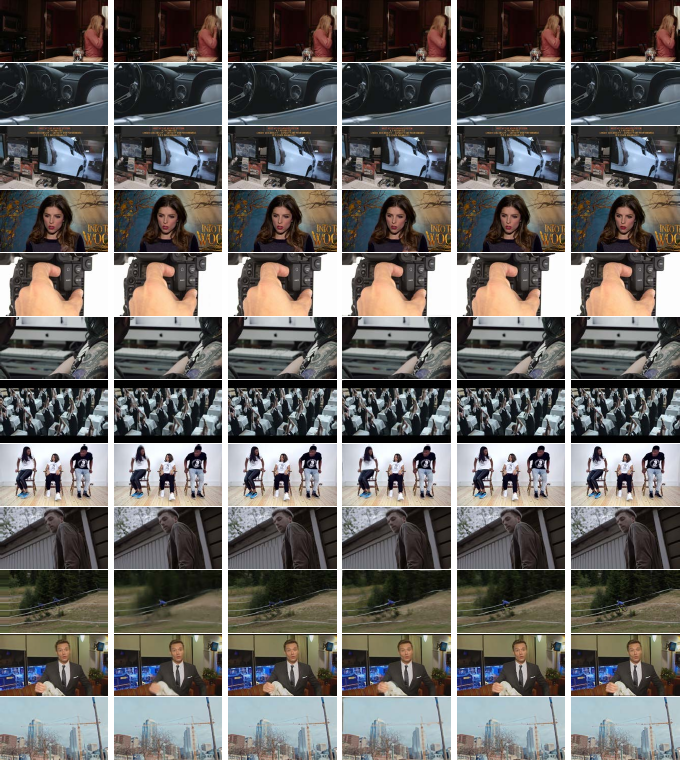}
    \setlength{\tabcolsep}{22pt}
    \begin{tabular}{cccccc}
        EpicFlow & AdaConv & \,\,SepConv & \!Fixed Flow & TOFlow & \!\!\!\!\!\! Ground Truth \\
    \end{tabular}
    \caption{Qualitative results on video interpolation. Samples are randomly selected from the Vimeo interpolation benchmark. The differences between different algorithms are clear only when zoomed in.}
    \label{fig:qualitativeinterp}
\end{figure*}

\begin{figure*}[p]
    \centering
    \includegraphics[width=0.9\linewidth]{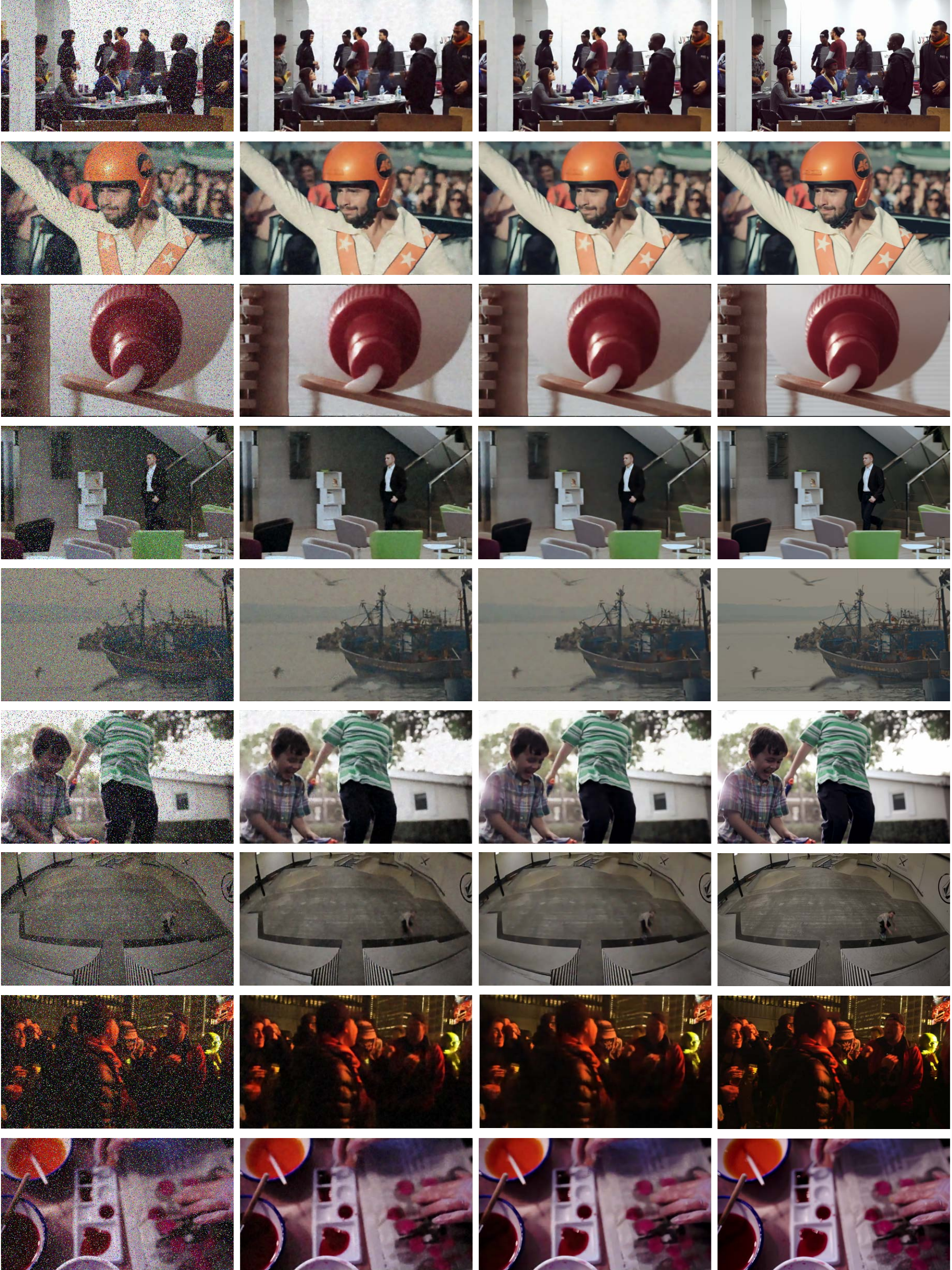}
    \begin{tabular}{C{4.2cm}C{3.2cm}C{3.2cm}C{4.2cm}}
         \,Input & \!\!Fixed Flow & \,\,\,\,\,\,\,TOFlow & \,\,\,Ground Truth\\
    \end{tabular}
    \caption{Qualitative results on RGB video denoising. Samples are randomly selected from the Vimeo denoising benchmark. The differences between different algorithms are clear only when zoomed in.}
    \label{fig:qualitativedenoisecolor}
\end{figure*}

\begin{figure*}[p]
    \centering
    \includegraphics[width=0.9\linewidth]{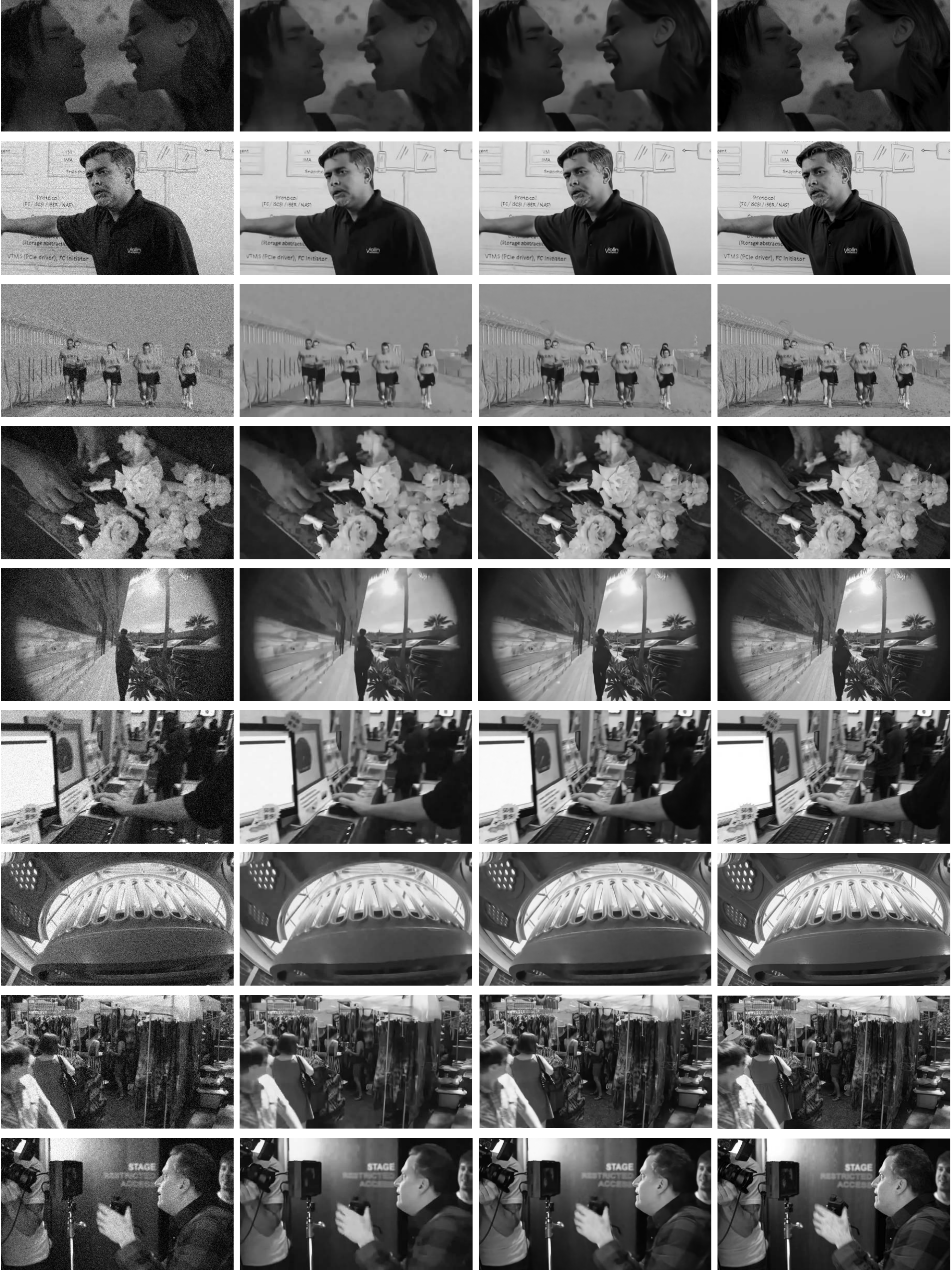}
    \begin{tabular}{C{4.2cm}C{3.2cm}C{3.2cm}C{4.2cm}}
         \,Input & \!V-BM4D & \,\,\,\,\,\,\,TOFlow & \,\,\,Ground Truth\\
    \end{tabular}
    \caption{Qualitative results on grayscale video denoising. Samples are randomly selected from the Vimeo denoising benchmark. The differences between different algorithms are clear only when zoomed in.}
    \label{fig:qualitativedenoisebw}
\end{figure*}

\begin{figure*}[p]
    \centering
    \includegraphics[width=0.9\linewidth]{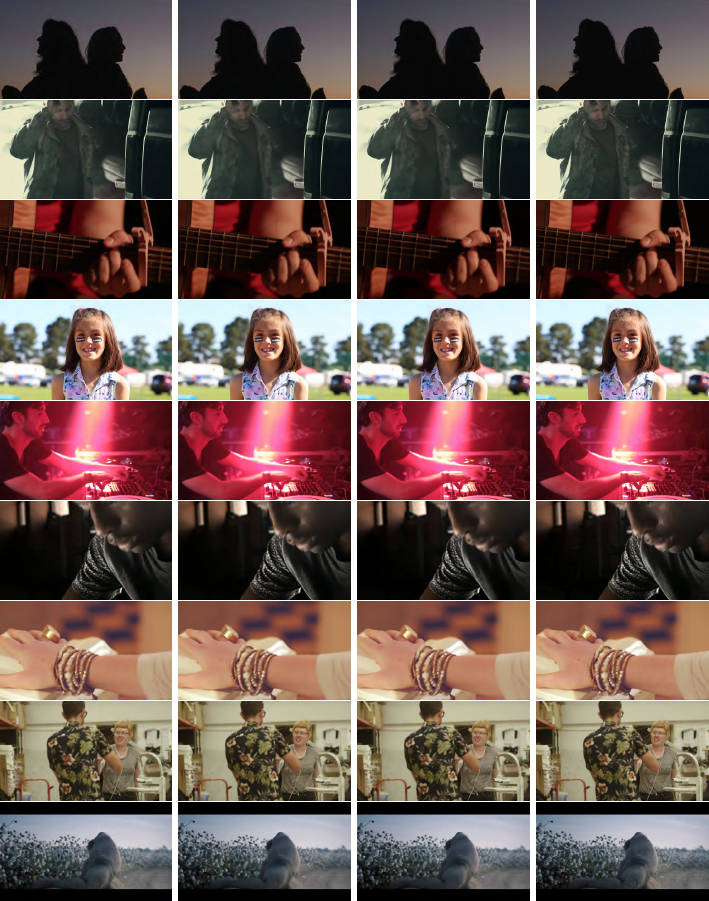}
    \begin{tabular}{C{4.2cm}C{3.2cm}C{3.2cm}C{4.2cm}}
         \,Input & \!V-BM4D & \,\,\,\,\,\,\,TOFlow & \,\,\,Ground Truth\\
    \end{tabular}
    \caption{Qualitative results on video deblocking. Samples are randomly selected from the Vimeo deblocking benchmark. The differences between different algorithms are clear only when zoomed in.}
    \label{fig:qualitativedeblock}
\end{figure*}

\begin{figure*}[p]
    \centering
    \includegraphics[width=0.98\linewidth]{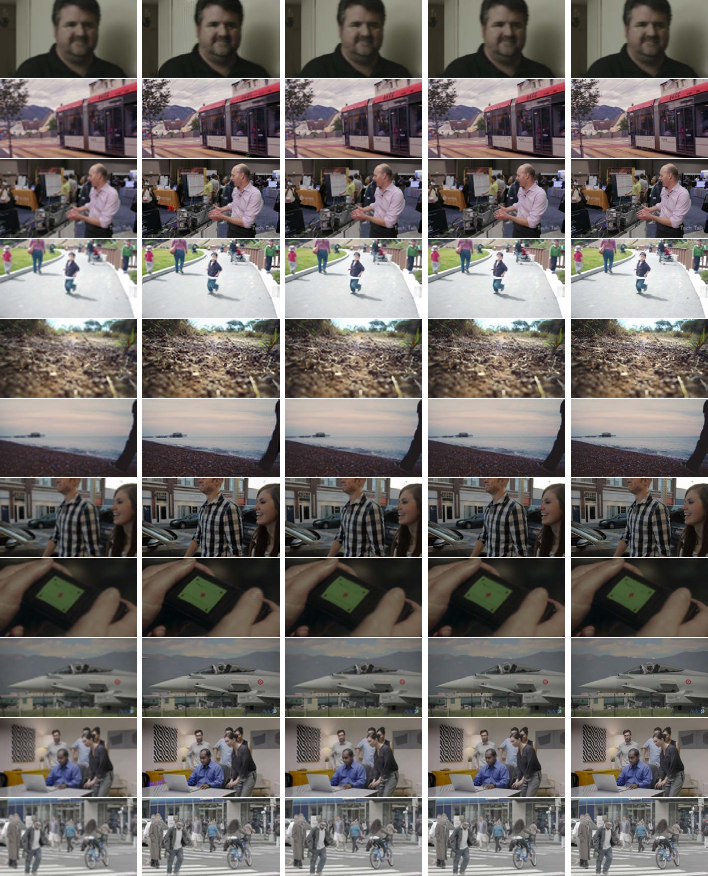}
    \setlength{\tabcolsep}{0pt}
    \begin{tabular}{C{3.42cm}C{3.42cm}C{3.42cm}C{3.42cm}C{3.42cm}}
        Bicubic & DeepSR & \,Fixed Flow & \,\,TOFlow & \,\,Ground Truth \\
    \end{tabular}
    \caption{Qualitative results on video super-resolution. Samples are randomly selected from the Vimeo super-resolution benchmark. The differences between different algorithms are clear only when zoomed in. DeepSR was originally trained on 30--50 images, but evaluated on 7 frames in this experiment, so there are some artifacts.}
    \label{fig:qualitativesr}
\end{figure*}

\myparagraph{Additional qualitative results. } We show additional results on the following benchmarks: Vimeo interpolation benchmark (\fig{fig:qualitativeinterp}), Vimeo denoising benchmark (\fig{fig:qualitativedenoisecolor} for RGB videos, and \fig{fig:qualitativedenoisebw} for grayscale videos), Vimeo deblocking benchmark (\fig{fig:qualitativedeblock}), and Vimeo super-resolution benchmark (\fig{fig:qualitativesr}). We randomly select testing images from test datasets. Differences between different algorithms are more clearer when zoomed in.

\myparagraph{Flow estimation module. } We use SpyNet~\citep{Ranjan2017Optical} as our flow estimation module. It consists of four sub-networks with the same network structure, but each sub-network has an independent set of parameters.  Each sub-network consists of five sets of 7$\times$7 convolutional (with zero padding), batch normalization and ReLU layers. The number of channels after each convolutional layer is 32, 64, 32, 16, and 2. The input motion to the first network is a zero motion field.

\myparagraph{Image processing module. } We use slight different structures in the image processing module for different tasks. For temporal frame interpolation both with and without masks, we build a residual network that consists of an averaging network and a residual network. The averaging network simply averages the two transformed frames (from frame 1 and frame 3).
The residual network also takes the two transformed frames as input, but calculates the difference between the actual second frame and the average of two transformed frames through a convolutional network consists of three convolutional layers, each of which is followed by a ReLU layer. The kernel sizes of three layers are 9$\times$9, 1$\times$1, and 1$\times$1 (with zero padding), and the numbers of output channels are 64, 64, and 3. The final output is the summation of the output of the averaging network and the residual network.

For video denoising/deblocking, the image processing module uses the same six-layer convolutional structure (three convolutional layers and three ReLU layers) as interpolation, but without the residual structure. We have also tried the residual structure for denoising/deblocking, but there is no significant improvement.

For video super-resolution, the image processing module 
consists of four pairs of convolutional layers and ReLU layers. The kernel sizes for these four layers are 9$\times$9, 9$\times$9, 1$\times$1, and 1$\times$1 (with zero padding), and the numbers of output channels are 64, 64, 64, and 3.

\myparagraph{Mask network. } Similar to our flow estimation module, our mask estimation network is also a four-level convolutional neural network pyramid as in \fig{fig:mask_net}. Each level consists of the same sub-network structure with five sets of 7$\times$7 convolutional (with zero padding), batch normalization and ReLU layers, but an independent set of parameters (output channels are 32, 64, 32, 16, and 2). For the first level, the input to the network is a concatenation of two estimated optical flow fields (four channels after concatenation), and the output is a concatenation of two estimated masks (one channel per mask). From the second level, the input to the network switch to a concatenation of, first, two estimated optical flow fields at that resolution, and second, bilinear-upsampled masks from the previous level (the resolution is twice of the previous level). In this way, the first level mask network estimates a rough mask, and the rest refines high frequency details of the mask. 

We use cycle consistencies to obtain the ground truth occlusion mask for pre-training the mask network. For two consecutive frames $I_1$ and $I_2$, we calculate the forward flow $v_{12}$ and the backward flow $v_{21}$ using the pre-trained flow network. Then, for each pixel $p$ in image $I_1$, we first map it to $I_2$ using $v_{12}$ and then map it back to $I_1$ using $v_{21}$. If it maps to a different point rather to $p$ (up to an error threshold of two pixels), then this point is considered to be occluded. 


\end{document}